\documentclass[letterpaper, 10 pt, journal, twoside]{IEEEtran}

\IEEEoverridecommandlockouts                              % This command is only needed if
\makeatletter
\let\NAT@parse\undefined
\makeatother
\usepackage[numbers]{natbib}

\usepackage[pdftex]{graphicx}
\usepackage{amsmath}
\usepackage{amssymb}  % assumes amsmath package installed
\usepackage{subfig}
\usepackage{multirow}
\usepackage{array,booktabs}
\usepackage{diagbox}
\usepackage{balance}
\usepackage[ruled]{algorithm}
\usepackage{algpseudocode}
 \usepackage[final]{hyperref}
 \hypersetup{
     colorlinks=true,
     linkcolor=blue,
     filecolor=magenta,
     urlcolor=blue,
     citecolor=black
 }
\usepackage[font=small]{caption}
\usepackage{./rpm_packages/rpm_SIunits}
\usepackage{./rpm_packages/rpm_acronyms}
\usepackage{./rpm_packages/rpm_math}
\usepackage{./rpm_packages/rpm_misc}

\usepackage{soul,color}
\usepackage{lipsum}

\usepackage{xcolor}
\usepackage[table,xcdraw]{xcolor}

\newcommand{\revision}[1]{{\textcolor{black}{#1}}}

\usepackage{tikz} % \checkmark

\title{XPRESS: X-band Radar Place Recognition\\ via Elliptical Scan Shaping}
\author{Hyesu Jang${}^{1}$, Wooseong Yang${}^{2}$, Ayoung Kim${}^{2}$, Dongje Lee${}^{1}$, and Hanguen Kim${}^{1*}$

\thanks{Manuscript received: July 1, 2025; Revised: September 10, 2025; Accepted: October 20, 2025. This paper was recommended for publication by Editor Giuseppe Loianno upon evaluation of the Associate Editor and Reviewers' comments.
This work was supported by the KIMST (07582969), Korea.}%Use only for final RAL version
\thanks{$^{1}$H. Jang, D. Lee, and H. Kim are with Seadronix Corp., Seoul, S. Korea {\tt\small [hsjang, dj.lee, hank05]@seadronix.com}}%
\thanks{$^{2}$W. Yang and A. Kim are with the Department of Mechanical Engineering, SNU, Seoul, S. Korea {\tt\small [yellowish, ayoungk]@snu.ac.kr}}%
}

\begin{document}

%\onecolumn
\maketitle

\begin{abstract}
X-band radar serves as the primary sensor on maritime vessels, however, its application in autonomous navigation has been limited due to low sensor resolution and insufficient information content. To enable X-band radar-only autonomous navigation in maritime environments, this paper proposes a place recognition algorithm specifically tailored for X-band radar, incorporating an object density-based rule for efficient candidate selection and intentional degradation of radar detections to achieve robust retrieval performance.
The proposed algorithm was evaluated on both public maritime radar datasets and our own collected dataset, and its performance was compared against state-of-the-art radar place recognition methods. An ablation study was conducted to assess the algorithm's performance sensitivity with respect to key parameters.
\end{abstract}
\section{Introduction}
\label{sec:intro}

% 1. Background & Why Maritime Place Recognition is Needed?
% \IEEEPARstart{M}{aritime} navigation has a long-standing history, evolving from manual control to modern autopilot systems. 
%The progression of vessels has been tightly coupled with advances in onboard hardware and sensing technologies. 
\IEEEPARstart{E}{arly} maritime autopilot systems were primarily designed for open-sea navigation, where the sparse and relatively unstructured environment allowed for sufficient autonomy despite intermittent sensor noise and signal fluctuations.
In contrast, near-shore, coastal, and port environments present more complex operational challenges. As demonstrated by \citet{han2019coastal} and \citet{jang2024moana}, \ac{GPS} signals in maritime environments are frequently subject to degradation and interference, complicating real-time decision-making in safety-critical scenarios. Additionally, these environments are characterized by high traffic density and dynamic obstacles, which complicate situational awareness and hinder robust localization due to the frequent occlusion and unpredictability of surrounding agents. Furthermore, geographic features shift over time under the influence of tidal effects and constructions. These challenges render the estimation of vessel location based solely on fixed \ac{ENC} or satellite images unreliable and necessitate the incorporation of real-time \ac{PR} with perception to account for dynamic environmental changes.
Previous studies \cite{sawada2023mapping} have utilized camera and \ac{LiDAR} sensors to perceive complex near-shore environments and enable autonomous sailing. However, their applicability in broader maritime scenarios is limited due to their short sensing range, which constrains path planning and vessel control. Moreover, these sensors are vulnerable to adverse weather conditions such as sea fog and squalls, and maintaining hardware functionality over the long term is challenging due to exposure to harsh environments and salt-induced corrosion.
%FIGURE
\begin{figure}[!t]
    \centering
    % \vspace{-2mm}
    \includegraphics[width=0.9\columnwidth]{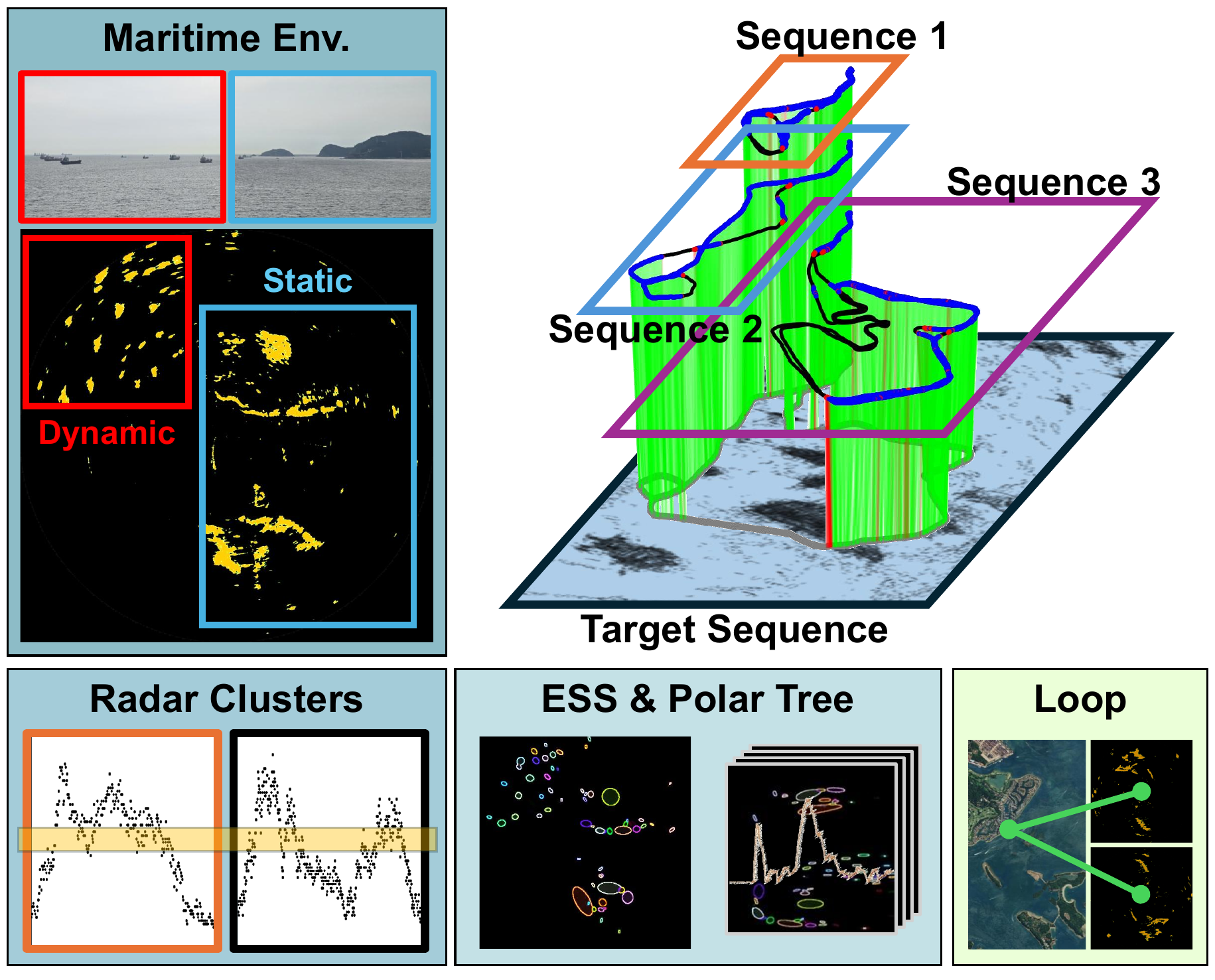}
    \caption{XPRESS proposes a fast and robust PR method for maritime environments using X-band radar. By leveraging radar cluster information and applying intentional elliptical degradation, the proposed approach minimizes the influence of dynamic elements and fluctuations of static objects.}
    \label{fig:intro}
    \vspace{-6mm}
\end{figure}
%FIGURE
% 2. What is the difference between ground and Maritime radar? & What are the challenges for x-band radar maritime place recognition?
Due to these complex factors, radar sensors are predominantly utilized in maritime environments, offering long-range detection capabilities and robust performance under adverse weather conditions. Notably, most existing vessels and yacht platforms are already equipped with X-band radar systems, the importance of developing algorithms tailored to the X-band radar is emphasized for the near-term deployment of autonomous vessel navigation.

The primary challenge lies in that X-band radar offers limited information while generating long-range image data, necessitating the extraction of salient details from low-resolution, large-scale inputs.
Furthermore, X-band radar possesses distinctive operational characteristics. \revision{Due to its low range resolution, lateral perturbations of the vessel have a negligible effect on the radar returns.} However, the system is particularly susceptible to rotational drift, a consequence of its low update rate and long-range detection configuration. Vessel rotation severely distorts the radar imagery, complicating the robust perception.
Therefore, algorithmic frameworks designed for X-band radar must prioritize and consider core data compression, rotational robustness, and computational efficiency.
%These considerations are essential for ensuring reliable perception and localization in large-scale maritime environments.
%where real-time operation and robustness are critical.

%FIGURE
\begin{figure*}[!t]
    \centering
    \includegraphics[width=0.7\textwidth]{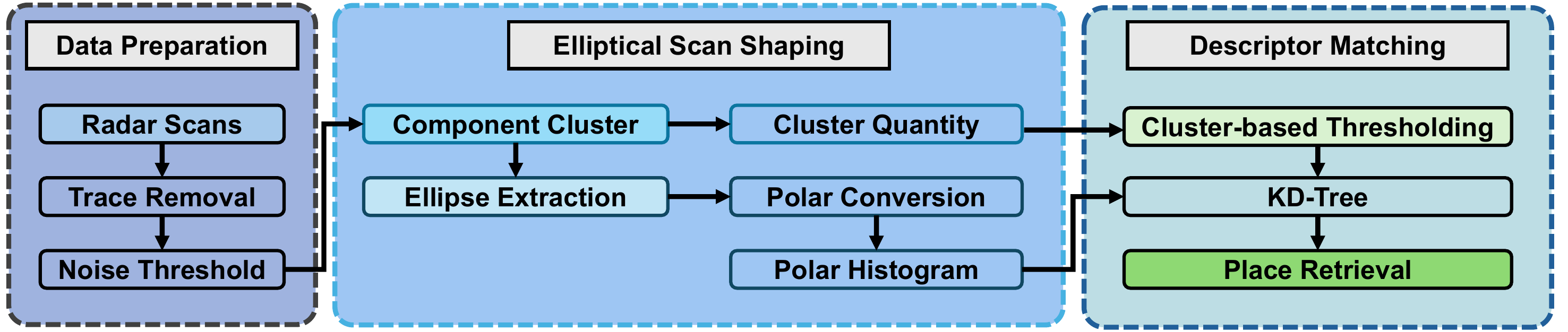}
    \caption{The concept of the proposed algorithm lies in the intentional degradation of radar data. Although static objects are detected, their appearance fluctuates across frames. To address this, radar data is first clustered using connected component labeling. Each labeled component is then approximated by an ellipse. The elliptical data is transformed into polar coordinates, from which a polar histogram is generated to enable KD-Tree-based retrieval. During descriptor matching, candidates are initially filtered based on the number of clusters, followed by estimation of revisitness among the selected candidates.}
    \label{fig:overview}
    \vspace{-6mm}
\end{figure*}
%FIGURE

% 3. What we have done?
We propose a \ac{PR} algorithm tailored for X-band radar, which resolves challenges due to the inherent characteristics of the sensor data. The main difficulties addressed by the algorithm include efficient extraction of salient information, reduction of retrieval latency, and achievement of rotational invariance. The specific contributions are summarized as follows:

\begin{itemize}
    \item \textbf{The first X-band radar \ac{PR} for maritime application}\\
    X-band radar captures abstract and sparse information, limiting to directly extract relevant features. To address this, we reinterpret the limitation as an opportunity to simplify the input representation, enabling the identification of common structural patterns across different observations. Based on the simplification, we design a descriptor specifically suited for \ac{PR} using X-band radar.
    
    \item \textbf{Fast, rotational-invariant, and fully-metric algorithm design for the huge and ambiguous x-band radar data}\\
    To extract candidates from the descriptor dictionary, we propose a two-step retrieval process. Since all radar frames are represented by clustered structures, we first estimate the number of clusters corresponding to each frame. Rather than directly comparing descriptor vectors, we initially apply an integer-based thresholding using the cluster count, reducing the search space. Subsequently, vector-wise matching is performed to identify the most appropriate candidate.
    
    \item \textbf{Robustness evaluation by structuring the ablation studies and own data acquisition}\\
    As the proposed work presents the first X-band radar-based \ac{PR} algorithm, thorough verification is essential. By benchmarking against existing state-of-the-art \ac{PR} algorithms originally developed for other scanning sensors, we demonstrate that our approach is specifically tailored for X-band radar and achieves robust performance in maritime environments.
\end{itemize}

\section{related work}
\label{sec:relatedwork}

% This paper is the first X-band radar-based \ac{PR} in maritime environments. In this section, we provide a brief introduction to prior works on W-band radar (also known as imaging radar)-based \ac{PR} and navigation in maritime environments.

\subsection{Place Recognition with Imaging Radar}
% imaging radar도 lidar 방법론 따라하다가 자기만의 특성을 고려한 방법으로 변화 : x-band도 자기만의 특성을 고려하는 방법론이 필요하다?
\revision{As noted in the radar survey by \citet{harlow2024new}, radars can be broadly categorized into imaging radars and automotive radars. X-band radar belongs to the imaging radar category, similar to W-band radar. The primary distinction lies in their operating bandwidths, with one utilizing the X-band and the other the W-band.}
% Both X-band and W-band radars are imaging radars that provide a top-down, two-dimensional image of the surrounding environment.
\revision{Due to the lack of PR method dedicated to X-band radar, we briefly summarize the previous PR work on W-band radar.
% In recent years, W-band radar has emerged as a promising sensor for ground autonomous vehicles due to its robust performance and extended range compared to \ac{LiDAR} and cameras, prompting advancements in radar-based \ac{PR}.
The early works on W-band radar have directly adopted the existing \ac{LiDAR}-based descriptor into radar imagery.}
\citet{kim2020mulran} exploited the Scan Context \cite{kim2018scan} representation by \ac{RCS} values for place retrieval.
Building on this, \citet{adolfsson2023tbv} integrated a Scan Context variant with the loop-verification module into TBV Radar SLAM.
\citet{hong2022radarslam} converted the radar image into point clouds and employed the M2DP descriptor for robust localization across diverse conditions.
\citet{jang2023raplace} utilized the Radon transform and the \ac{FFT} to create roto-translationally invariant descriptors, RaPlace.
Recent approaches have started to consider radar-specific characteristics. \citet{gadd2024open} combined feature aggregation with \ac{FFT} on radar imagery, 
\citet{kim2024referee} leveraged the distribution of unoccupied pixels in radar images as an informative descriptor, named Referee.
Learning-based methods have also been proposed for \ac{PR} with  W-band radar. \citet{barnes2020under} proposed self-supervised keypoint detection for global descriptors, and \citet{suaftescu2020kidnapped} embedded radar imaginaries into a metric feature space for similarity comparisons. \citet{yuan2023off} introduced uncertainty-aware metric learning to enhance robustness against environmental variations.
However, these learning-based approaches exhibit limited generalizability across varying radar noise characteristics in diverse environments. To tackle this issue, some approaches exploited consistent quality \ac{LiDAR}-derived maps to supervise radar embeddings \cite{yin2021rall, nayak2024ralf}.

% * While W-band radar’s dense, high-resolution returns are effectively utilized in ground vehicle scenarios, X-band radar, as explored in the MOANA framework, presents distinct challenges for maritime environments due to its coarser angular resolution and altered propagation behavior over water surfaces. These differences make direct application of W-band radar-based place recognition (PR) methods suboptimal for maritime settings. Consequently, a specialized PR methodology tailored to the unique signal characteristics and scattering properties of X-band radar is essential for achieving reliable localization on the open sea.
As noted by \citet{jang2024moana}, X-band radar has distinct data characteristics against W-band radar, prompting the necessity for specialized \ac{PR} methodology.
We incorporate the unique signal and data properties of the X-band radar to build novel descriptors, enabling reliable open-sea localization.

% 실험을 통해서 imaging radar pr방법이 잘 안되는걸 보였다는 내용 추가?

\subsection{Maritime SLAM and Perception}

The advancements in maritime navigation have largely focused on improving localization accuracy using marine radar, particularly X-band radar, for odometry purposes. For example, \citet{bibby2010hybrid} proposed a hybrid \ac{SLAM} framework that integrates both spatial and temporal data, enabling robust localization in dynamic marine environments. \citet{han2019coastal} introduced Coastal \ac{SLAM}, which combines marine radar with coastal area modeling for \ac{GPS}-restricted conditions. \citet{schiller2022improving} improved marine radar odometry by modeling radar resolution and incorporating temporal information. \citet{jang2024lodestar} introduced LodeStar, a novel method specifically designed for X-band radar odometry. \revision{However, there is a lack of PR methods using marine radar that can support the operation of a full SLAM system.}
Recent maritime \ac{SLAM} attempted to exploit multi-sensor fusion. Some efforts have integrated \ac{LiDAR} to enhance berthing state estimation and assistance systems \cite{hu2022estimation, wang2024berthing}, thereby improving situational awareness during complex maneuvers. Furthermore, \citet{sawada2023mapping} demonstrated \ac{LiDAR}-based \ac{SLAM} for coastal mapping and localization in GNSS-denied environments. \citet{lu2025multi} have further improved berthing navigation by fusing \ac{IMU} and RTK data. However, these solutions are primarily applicable to coastal and near-shore environments, where \ac{LiDAR} can be effective. In contrast, using \ac{LiDAR} in open-sea navigation proves to be impractical, as highlighted by \citet{jang2024moana}.
Therefore, we focus on exploring \ac{PR} methods that leverage the distinctive features of X-band radar to address the gaps in maritime \ac{SLAM} systems in wide oceanic areas.

% We explore the noticeable gap in the exploration of place recognition and \ac{SLAM} systems utilizing X-band radar, which remains underdeveloped.

% The challenges in using \ac{LiDAR} for open-sea navigation underscore the need for dedicated place recognition methodologies explicitly tailored to the unique characteristics of X-band radar. Unlike \ac{LiDAR}, X-band radar offers robust detection capabilities in vast, open-water environments where other sensors like \ac{LiDAR} and GPS are inadequate. Consequently, the development of place recognition systems for X-band radar is critical for improving the performance of maritime localization, particularly in long-range, autonomous, and continuous navigation scenarios. 

\section{X-band Radar Place Recognition}
\label{sec:method}
%This section details the design of the descriptor for X-band radar-based place recognition, the development of a fast and robust retrieval algorithm, and the extended application within a \ac{SLAM} framework.
\revision{X-band radar provides abstract representations of the surrounding environment, demonstrating challenges in navigation algorithms. The most challenging phenomenon is the inconsistency of detection details, even when the vessel remains in the same position.
Furthermore, it is necessary to account for large and clustered dynamic objects anchored in designated zones, potentially appear as dynamic islands in radar imagery.
The overall framework of the proposed algorithm consists of two stages, where the first involves generating an X-band radar descriptor that suppresses the effects of fluctuations and clustered dynamic objects, and the second performs efficient descriptor matching with predefined descriptor data. The overview of our system is depicted in \figref{fig:overview}}

\subsection{Marine Descriptor}
% Enabling \ac{PR} with X-band radar primarily requires effective extraction of relevant features while minimizing the influence of large-scale dynamic objects. Addressing these two challenges, we construct the descriptor as follows.
%

\begin{figure}[!t]
    \centering
    \includegraphics[width=\columnwidth]{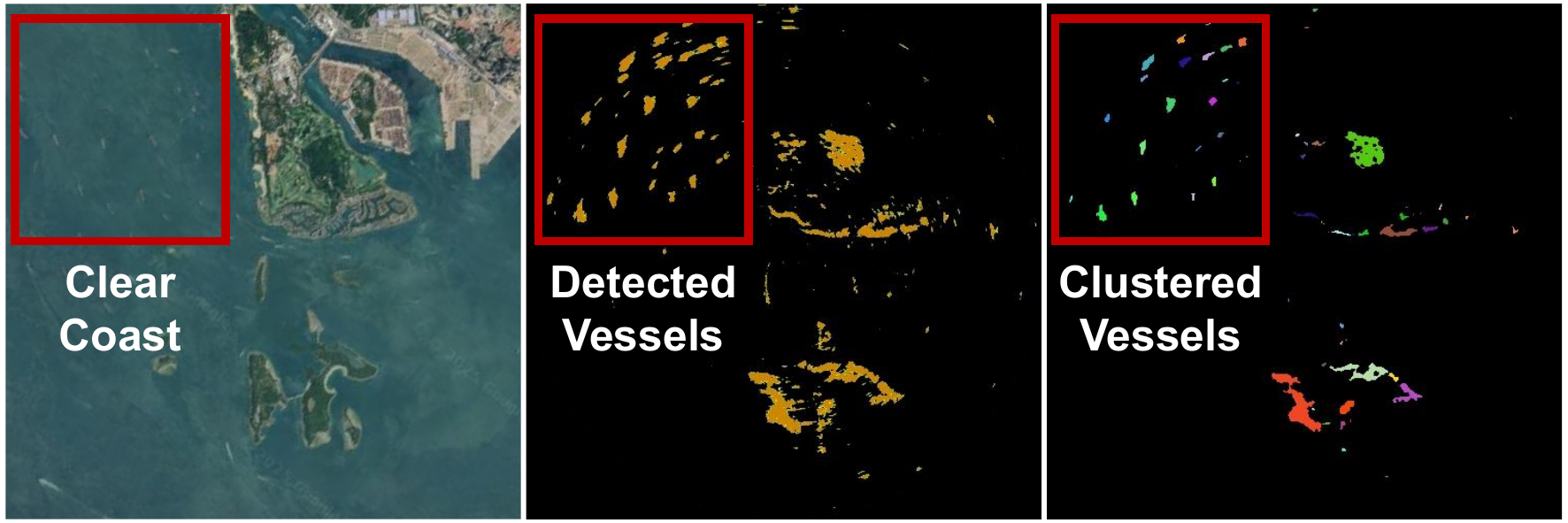}
    \vspace{-5mm}
    \caption{In coastal areas, while the density of anchored vessels remains relatively stable, their individual positions vary consistently. Considering the vessel density within specific regions, rather than relying on exact vessel positions, is advantageous for handling dynamic elements in X-band radar images.}
    \label{fig:CCL}
    \vspace{-6mm}
\end{figure}

\subsubsection{Connected Component Labeling (CCL)}
LiDAR and W-band radar can capture detailed structural information from the surrounding environment, enabling reliable estimation of surface planes or linear features. In contrast, X-band radar lacks the resolution necessary to visualize precise shapes of detected objects. Therefore, we treat radar returns as aggregated clusters of reflections. To efficiently segment these clusters, we employ CCL~\cite{samet1988component}. The nearby pixels $N(p)$ for the individual pixel $p$ are defined as
\begin{equation}
\centering
\begin{split}
N(p) &= \{(x\pm1,y),(x,y\pm1),(x\pm1,y\pm1)\} \\
N_{half}(p) &= \{(x\pm1,y-1),(x,y-1),(x-1,y)\}.
\end{split}
\label{eq:nearby}
\end{equation}
CCL is performed in two phases. In the first phase, the algorithm scans the radar image in a row-wise manner and assigns a label $l$ to each non-zero pixel. For each pixel $p$, if all neighboring pixels in the half-neighborhood $N_{half}(p)$ are zero, a new label is assigned to $p$. If there is exactly one non-zero neighbor, $p$ inherits the same label as that neighbor. If multiple non-zero neighbors exist with different labels, $p$ is assigned one of those labels arbitrarily, and the differing label pairs are stored as equivalence pairs, indicating that they belong to the same connected region.

In the second phase, once the initial labeling is complete, the equivalence \revision{pair} information is used to rearrange and unify the labels. Each group of equivalent labels is replaced with a unique representative label, ensuring that all pixels within the same connected component are assigned the same label. As a result, distinct clusters in the radar image are identified with unique label values.

\subsubsection{Elliptical Scan Shaping}
Exploiting the segmented clusters obtained from CCL, we derive elliptical scan shapes to simplify the representation of detected regions. Given the frame-to-frame fluctuations in radar detection, approximating each cluster with a parametric elliptical model improves the robustness of the descriptor by providing a consistent and compact geometric abstraction. The ellipse is defined by five parameters: the centroid $(x_c,y_c)$, the lengths of the major and minor axes $a$ and $b$, and the orientation $\theta$ of the major axis with respect to the reference coordinate frame.
We compute the centroid $(x^l_c,y^l_c)$ of cluster $l$ as
\begin{equation}
\centering
(x^l_c,y^l_c) = \left(\frac{1}{N_l}\sum_{i=1}^{N_l}x^l_i, \frac{1}{N_l}\sum_{i=1}^{N_l}y^l_i \right)
\label{eq:centroid}
\end{equation}
where $(x^l_i,y^l_i)$ are the coordinates of the $i$-th pixel within a cluster $l$, and $N_l$ is the total number of pixels in cluster $l$.
To determine the best-fit ellipse for each cluster, we compute the covariance matrix of the pixel coordinates within the cluster as
\begin{equation}
\centering
\vspace{-2mm}
\begin{split}
C^l &= \begin{bmatrix}
C^{l}_{11} & C^{l}_{12} \\
C^{l}_{21} & C^{l}_{22}
\end{bmatrix}\\
&= \frac{1}{N_l}\begin{bmatrix}
\sum(x^l_i-x^l_c)^2 & \sum(x^l_i-x^l_c)(y^l_i-y^l_c) \\
\sum(x^l_i-x^l_c)(y^l_i-y^l_c) & \sum(y^l_i-y^l_c)^2
\end{bmatrix}
\end{split}
\label{eq:cov}
\end{equation}
By performing eigen decomposition on the covariance matrix $C^l$, we obtain the eigenvalues $\lambda_{1,2}$ with
\begin{equation}
\centering
\lambda_{1,2} = \frac{1}{2}\left((C^{l}_{11}+C^{l}_{22})\pm\sqrt{(C^{l}_{11}-C^{l}_{22})^2+4(C^{l}_{12})^2}\right).
\label{eq:centroid}
\end{equation}

Assuming a Gaussian pixel distribution, the lengths of the major and minor axes of the ellipse are defined as $a = 2\sqrt{\lambda_1}$ and $b = 2\sqrt{\lambda_2}$, respectively.
To estimate the orientation of the ellipse, we use the components of the covariance matrix,
\begin{equation}
\centering
\theta^l = \frac{1}{2}\arctan\left(\frac{2C^{l}_{12}}{C^{l}_{11}-C^{l}_{22}}\right).
\label{eq:centroid}
\end{equation}
%The resulting centroid serves as a representative feature point for the corresponding radar cluster.
We utilize the elliptical scan-shaped image $I_{cart}$ as initial x-band radar descriptor, which suppresses irregular fluctuations and enhances the representation of static regions. For all clusters, elliptical scan-shaped images are defined as
\begin{equation}
\centering
\begin{split}
x^l(t) = x^l_c + a^l \cos{(t)}\cos{(\theta^l)} - b^l \sin{(t)}\sin{(\theta^l)}\\
y^l(t) = y^l_c + a^l \cos{(t)}\sin{(\theta^l)} + b^l \sin{(t)}\cos{(\theta^l)}
\end{split},
\label{eq:cand1}
\end{equation}
\begin{equation}
\centering
I_{cart} = \bigcup_{l=1}^{l_F} \left\{ \left( x^l(t), y^l(t) \right) \; \Big| \; t \in [0, 2\pi] \right\}.
\label{eq:cand1}
\end{equation}

\subsection{Maritime Place Recognition}
The minimum detection range of X-band radar in the MOANA dataset is about $75\,m$ and the range resolution is about $3.25\,m/px$, making the algorithm unsuitable for focusing on \revision{lateral movement and invariance}. Effective \ac{PR} algorithms require both fast retrieval and sensor-specific descriptor. Therefore, our algorithm emphasizes the descriptor including rotational invariance while maintaining high retrieval speed.

\subsubsection{Cluster Searching}
To minimize the time complexity of the search algorithm, we propose a fast and reliable initial candidate extraction. Maritime scenes captured by X-band radar consist of both static and dynamic objects. Although dynamic objects are fewer in number, their physical size is often substantial which can be misinterpreted as small static islands. This characteristic limits the applicability of conventional feature-based or direct matching techniques for robust place retrieval. Furthermore, as illustrated in \figref{fig:CCL}, numerous vessels are at anchored in anchorage, which subsequently act as dynamic objects. For long-time-interval \ac{PR}, it is essential to recognize the presence of anchoring zones, while accounting for the characteristic that the positions of individual vessels within the anchoring zones differ from the previous data acquisition. To address this, we treat dynamic objects as individual clusters and consider only the number of clusters, irrespective of their size. This resolves the challenges from both large-scale dynamic objects and anchoring zones, where the configuration of vessels varies over time, yet the overall vessel density remains relatively consistent.

A frame $F$ is included in the candidate set $F_{\textrm{cand}}$ if and only if the difference in cluster count between $F$ and the query frame $Q$ satisfies the threshold condition,
\begin{equation}
\centering
F \in F_{\textrm{cand}} \Longleftrightarrow \abs{l_F - l_Q} \leq p
\label{eq:cand1}
\end{equation}
$l_F$ and $l_Q$ denote the number of clusters in the candidate frame and the query frame, respectively, and $p$ is the predefined threshold.
Compared to vector-to-vector comparisons, thresholding based on a single integer accelerates the candidate selection process.

%Using this standard and the buffer value, we can approximately extract the candidate place where the query data belongs.

\begin{figure}[!t]
    \centering
    \includegraphics[width=0.85\columnwidth]{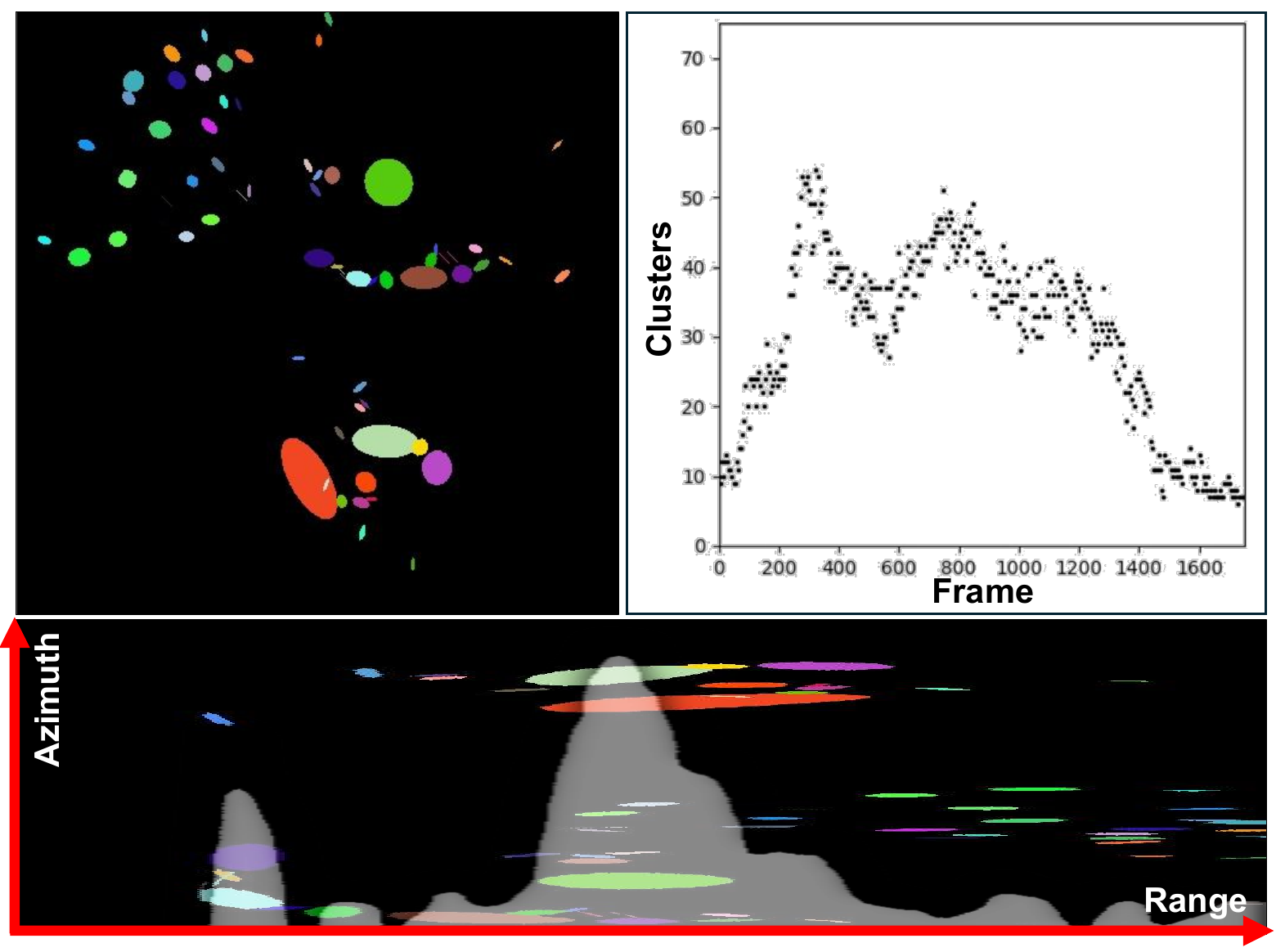}
    \label{fig:ellipses}
    % \vspace{-5mm}
    \caption{X-band radar data exhibits fluctuations, the same objects across consecutive frames appear with varying cluster boundaries. To mitigate the impact of this variability, the radar data is degraded into elliptical representations (top-left). The number of resulting clusters (top-right) serves as a criterion for initial candidate selection, and the histogram derived from the polar-transformed data is used as a descriptor for location retrieval (bottom).}
    \vspace{-6mm}
\end{figure}

\subsubsection{Polar Ellipse Matching}
After extracting initial candidates, we perform a rotationally invariant search by converting the radar data into polar coordinates. 
With the central point $(x_0,y_0)$ of the Cartesian image, the polar coordinate transformation is performed by first translating the points as $\tilde{x}(t) = x(t) - x_0$, $\tilde{y}(t) = y(t) - y_0 $, and then rearranging them into range-azimuth coordinates, where
\begin{equation}
\centering
r(t),\phi(t) = \sqrt{ \tilde{x}(t)^2 + \tilde{y}(t)^2 },\; \arctan\left( \frac{\tilde{y}(t)}{\tilde{x}(t)} \right)
\label{eq:cand1}
\vspace{-4mm}
\end{equation}

\begin{equation}
\centering
I_{\textrm{polar}} = \bigcup_{l=1}^{l_F} \left\{ \left( r^l(t), \phi^l(t) \right) \; \Big| \; t \in [0, 2\pi] \right\}
\label{eq:polar}
\end{equation}

We integrate pixel intensities along each angular direction to generate a range-wise histogram as
\begin{equation}
\centering
V_{\textrm{ESS}} = \sum_{\phi} I_{\textrm{polar}}
\label{eq:ess}
\end{equation}
that incorporates rotational invariance. This integration also mitigates uncertainty arising from anchoring zones, where dynamic vessels are randomly distributed.
Exploiting a KD-tree structure, we efficiently retrieve the nearest neighbor vector among the candidates, and verify the final match by evaluating the cosine similarity distance.

% \begin{equation}
% \centering
% d(u,v) = 1-\frac{ \mathbf{u} \cdot \mathbf{v} }{ \| \mathbf{u} \| \, \| \mathbf{v} \| }
% \label{eq:cossim}
% \end{equation}
\section{experiment}
\label{sec:experiment}
\subsection{Maritime Dataset Configuration}
\begin{figure}[!t]
    \centering
    \includegraphics[width=\columnwidth]{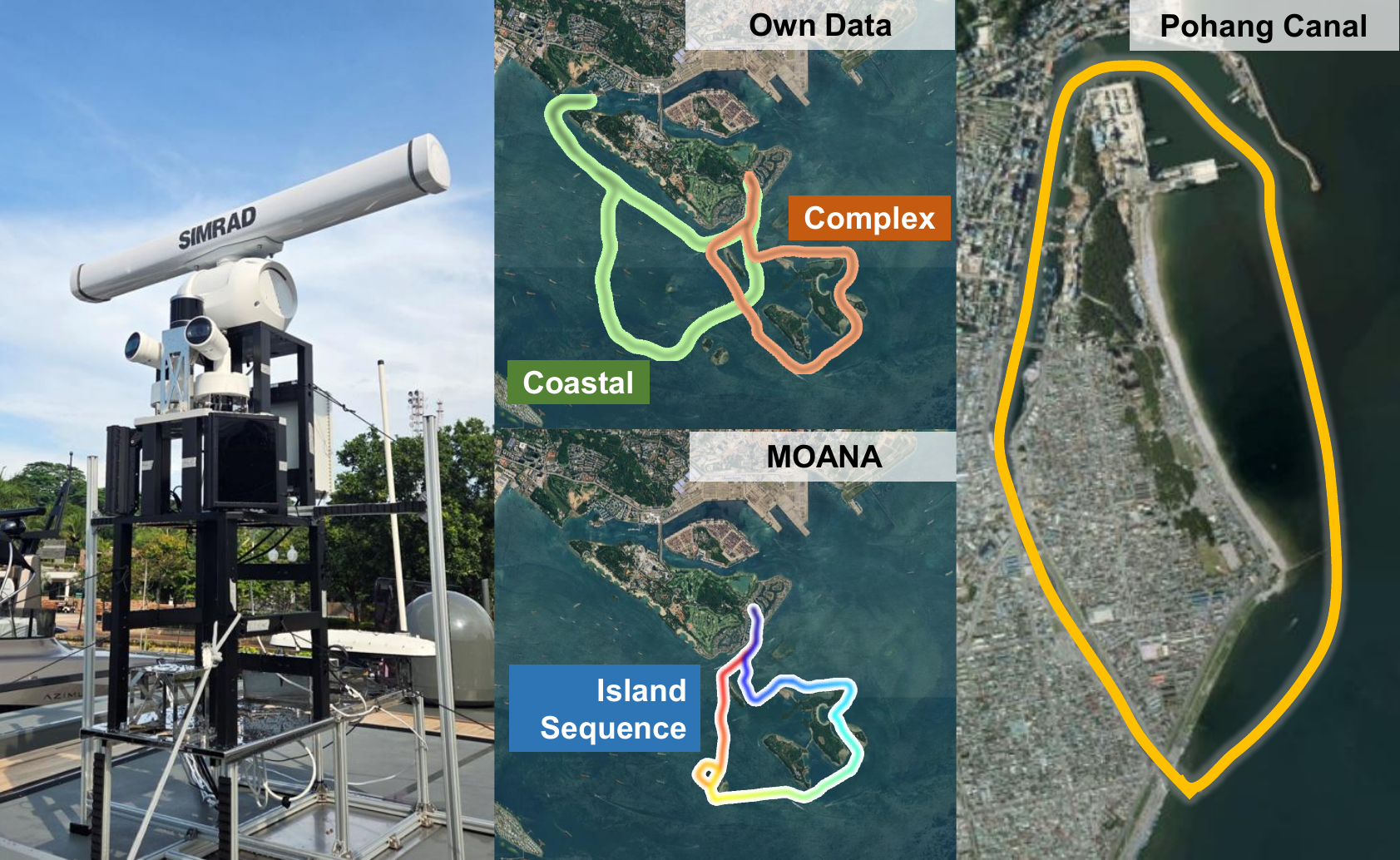}
    \caption{The left image illustrates our sensor set and the rights are the routes for the evalutation. To evaluate the proposed algorithm, we utilized two public datasets and additionally acquired our own dataset in the same geographic region as the MOANA dataset.  Our dataset is divided into two sequences, Coastal and Complex, which are used to assess intra-session and long-term inter-session \ac{PR}, respectively.}
    \label{fig:dataset}
    \vspace{-6mm}
\end{figure}

To validate the robustness of our algorithm, we conducted experiments on two major public maritime datasets as well as our own collected dataset. The MOANA dataset~\cite{jang2024moana} contains diverse routes with multiple loops, suitable for evaluating both intra-session loop closure and inter-session \ac{PR} performance. In contrast, the Pohang Canal Dataset~\cite{chung2023pohang} does not include intra-session loops but provides repeated traversals of the same routes across different sequences, which is appropriate for assessing inter-session \ac{PR} capabilities. Additionally, to demonstrate the generalized performance across different datasets, we collected a new dataset covering routes adjacent to the MOANA dataset, enabling the robustness evaluation of inter-dataset \ac{PR}. The maximum detection range of the radar in our dataset is $3328\,m$, with a identical pixel resolution to the MOANA. The detailed experimental environment settings are summarized in \figref{fig:dataset}.

\subsubsection{MOANA dataset}
The MOANA dataset consists of seven sequences designed to evaluate maritime navigation algorithms. Among these, the Island sequences are particularly suited for \ac{PR} evaluation. We use three Island sequences for intra-session \ac{PR} and the Island Expedition sequence for inter-session \ac{PR} experiments.
The Single Island sequence provides clear radar data, enabling the evaluation of short-term revisitness detection. The Twins Island sequence offers noisier radar data, challenging the robustness of \ac{PR}. The Complex Island sequence combines both clear and noisy radar data, with a large number of loop candidates due to its longer navigation route.
The Island Expedition sequence encompasses all locations from the Single, Twins, and Complex Island sequences. Thus, we present inter-session \ac{PR} results across three pairs: Single-Expedition, Twins-Expedition, and Complex-Expedition.

\subsubsection{Pohang Canal dataset}
Since the Pohang Canal dataset consists of a single route without loops, we evaluated inter-session recognition performance. The Narrow Canal area represents the most challenging environment for X-band radar, characterized by significant multipath noise and frequent detection failures. The Inner Port area, being wider than the Narrow Canal, is suitable for evaluating tunnel-like \ac{PR} scenarios where similar features appear repeatedly. The Outer Port area, although relatively short, connects the narrow region to the broader coastal area. The Near-Coastal area provides stable radar operation conditions. We validated our algorithm across these four distinct radar environments at once.

\subsubsection{Own dataset}
Our dataset was collected in the same maritime environment as the MOANA dataset, sharing common areas at several locations. Using this dataset, we evaluate both intra-session and long-time-interval inter-session \ac{PR} by matching against the Island Expedition sequence from the MOANA dataset. This experiment verifies the robustness of the proposed algorithm when applied to independently collected X-band radar data.

\subsection{Evaluation Criteria}
\ac{PR} in the context of SLAM differs fundamentally from general image retrieval tasks. In SLAM, the objective is to identify the single most probable frame from the previous data to enable loop closure or designate a keyframe. Additionally, the system performance of processing large-scale data must be considered to ensure real-time operation. Taking these requirements into account, we adopt the following criteria for evaluating the main results.
\subsubsection{\ac{PR} Methods}
As introduced in the previous chapter, four recent W-band radar-based \ac{PR} algorithms are available for comparison\cite{kim2020mulran}\cite{jang2023raplace}\cite{gadd2024open}\cite{kim2024referee}. However, we excluded RaPlace from our performance evaluation due to its high computational complexity, which does not align with our objective of assessing the efficiency and lightness of the algorithm. All algorithms were reproduced using the W-band radar data to verify their integrity, and the results confirmed that performance was not degraded on the W-band radar dataset.
% Radar scan context\cite{kim2020mulran}
% Referee\cite{kim2024referee} is the most powerful place recogntion method for the w-band radar, that estimate the places utilizing the free spaces that minimizes the effects of the false detections. The concept X-band radar

\subsubsection{Precision-recall curve and F1 curve}
The precision-recall curve and the area under the curve (AUC) are standard metrics for evaluating \ac{PR} accuracy. These indices measure the model's positive predictive value ($\textrm{=TP/(TP+FP)}$) and its ability to correctly identify true matches ($\textrm{=TP/(TP+FN)}$).
% \begin{equation}
% \centering
% Precision = \frac{TP}{TP+FP}, Recall = \frac{TP}{TP+FN}
% \label{eq:FAR}
% \end{equation}
The F1 score is the harmonic mean of precision and recall, effectively balancing the two metrics to minimize bias toward either. The F1 curve is plotted against recall values, sharing the same x-axis as the precision-recall curve.
% \begin{equation}
% \centering
% \text{F1} = \frac{2\cdot\text{Precision\cdot Recall}}{\text{Precision}+\text{Recall}}
% \label{eq:F1}
% \end{equation}

\subsubsection{False Alarm Rate and Recall@1}
The false alarm rate ($\textrm{=FP/(TN+FP)}$) serves as a critical metric to assess whether the algorithm reliably identifies only true positive matches. In maritime environments where navigation occurs over expansive areas without fixed lanes and loop closures are rare, false alarms can severely impact system reliability, even if precision is high. 
% The false alarm rate is defined as 
% \begin{equation}
% \centering
% \textrm{FAR} = \frac{\textrm{FP}}{\textrm{TN}+\textrm{FP}}
% \label{eq:FAR}
% \end{equation}
Recall@1 $\textrm{(=TP/GT)}$ is not a fully appropriate metric for SLAM applications, as it assumes the presence of at least one correct match among the candidates, whereas SLAM systems may encounter cases where no true correspondence exists. 
The false alarm rate necessitates a reasonably performing classifier that guarantees the true positive rate, and recall@1 is an unsafe index since the false alarms are not considered.
Both the false alarm rate and Recall@1 require method-specific optimal thresholds to be effectively utilized as evaluation metrics. Therefore, these two factors were jointly considered to ensure a fair and reliable assessment of model performance.
For intra-session \ac{PR}, we employ the area under the \ac{ROC} curve to jointly evaluate the false alarm rate and recall. For inter-session \ac{PR}, where correct matches are known to exist among the candidates, we adopt Recall@1 as the evaluation metric.
% Nevertheless, we include Recall@1 as an auxiliary metric to evaluate the pure \ac{PR} capability of the algorithm.

\subsection{Intra-session Place Recognition}
For efficient and generalizable evaluation, the algorithm was assessed at every five frame interval, with a revisit criterion defined as a $100\,m$ range, which corresponds to approximately three percent of the radar maximum detection range. All experiments were performed on a system equipped with an Intel® Core™ Ultra 7 165H CPU and 32 GB of RAM.

\begin{figure}[!t]
    \centering
    \includegraphics[width=0.95\columnwidth]{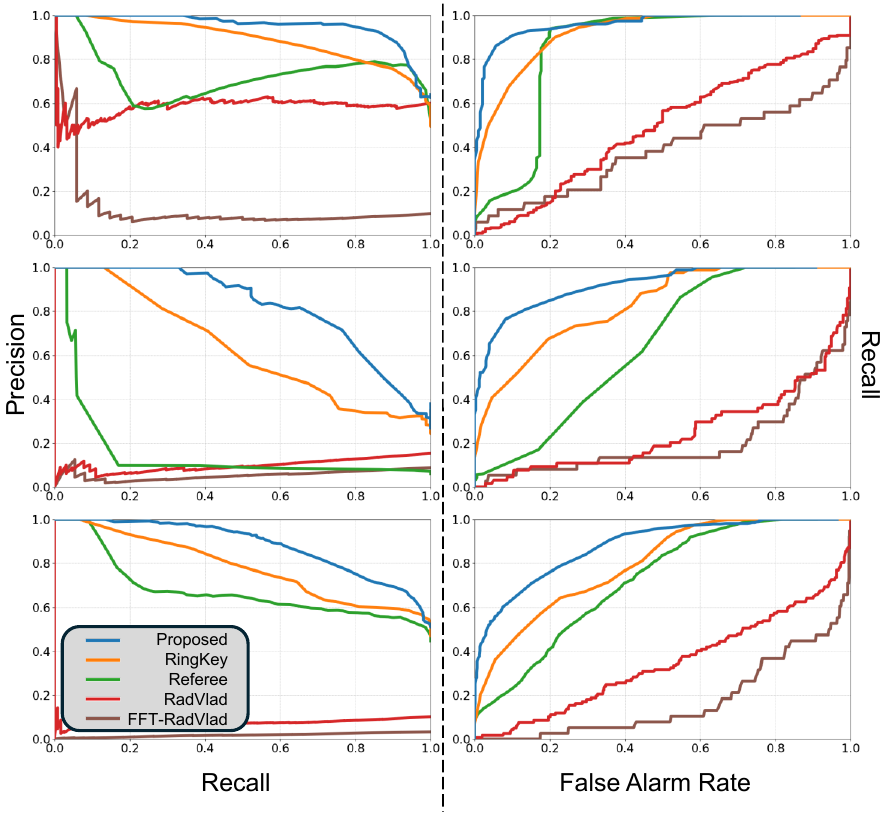}
    \caption{Precision-recall and ROC curves for intra-session \ac{PR} on the MOANA dataset (from top to bottom: Single, Twins, Complex) demonstrate that the proposed algorithm achieves the most robust performance among state-of-the-art radar \ac{PR} methods.}
    \label{fig:curv_result}
    \vspace{-6mm}
\end{figure}

% \subsubsection{MOANA Dataset}
As shown in \tabref{tab:pr_result} and \figref{fig:curv_result}, our algorithm consistently outperforms across all sequences. Notably, in the Twins Island sequence, Referee\cite{kim2024referee} exhibits comparatively poor performance. This can be attributed to the characteristics of X-band radar in open ocean settings. While Referee is capable of generating robust descriptors in structured and complex environments by leveraging free space regions in the image, its discriminative capability degrades when free space becomes dominant. In this sequence, Referee produces a high number of false alarms in areas lacking distinctive detections.

RadVLAD and FFT-RadVLAD are effective algorithms for W-band radar; however, they exhibited performance degradation when applied to X-band radar. The primary suspected reason is the limited information in X-band radar images, where the simpler pixel composition makes it difficult for the VLAD descriptor to distinguish between different places.
% \subsubsection{Own Dataset}

% \usepackage{multirow}
\begin{table}[]
\centering
\resizebox{\columnwidth}{!}{
\begin{tabular}{ccccccc}
\hline
\multicolumn{2}{c}{Sequence}     & Method      & Time(ms)         & AUC         & F1        & ROC                      \\ \hline
\multicolumn{1}{c|}{}        & \multicolumn{1}{c|}{}   & \multicolumn{1}{c|}{RingKey~\cite{kim2018scan}}     &    0.08    &   \underline{0.89}   &   0.82   &   \underline{0.92}  \\
\multicolumn{1}{c|}{}        & \multicolumn{1}{c|}{}   & \multicolumn{1}{c|}{Referee~\cite{kim2024referee}}     &    \underline{0.07}    &   0.73   &   \underline{0.85}   &   0.85  \\
\multicolumn{1}{c|}{}        & \multicolumn{1}{c|}{}   & \multicolumn{1}{c|}{RadVLAD~\cite{gadd2024open}}     &    9.23    &   0.59   &   0.75   &   0.49  \\
\multicolumn{1}{c|}{}        & \multicolumn{1}{c|}{}   & \multicolumn{1}{c|}{FFT-RadVLAD~\cite{gadd2024open}} &    9.46    &   0.12   &   0.18   &   0.38  \\
\multicolumn{1}{c|}{}        & \multicolumn{1}{c|}{\multirow{-5}{*}{\begin{tabular}[c]{@{}c@{}}Single\\ Island\end{tabular}}}  & \multicolumn{1}{c|}{\cellcolor[HTML]{EFEFEF}Ours} & \cellcolor[HTML]{C4E1E6}\textbf{0.04}    & \cellcolor[HTML]{C4E1E6}\textbf{0.95}   & \cellcolor[HTML]{C4E1E6}\textbf{0.90}   & \cellcolor[HTML]{C4E1E6}\textbf{0.96}    \\ \cline{2-7} 
\multicolumn{1}{c|}{}        & \multicolumn{1}{c|}{}   & \multicolumn{1}{c|}{RingKey}     &   0.09   &   \underline{0.63}   &   \underline{0.56}   &    \underline{0.83}  \\
\multicolumn{1}{c|}{}        & \multicolumn{1}{c|}{}   & \multicolumn{1}{c|}{Referee}     &   \underline{0.08}   &   0.15   &   0.16   &    0.65  \\
\multicolumn{1}{c|}{}        & \multicolumn{1}{c|}{}   & \multicolumn{1}{c|}{RadVLAD}     &   11.27  &   0.10   &   0.27   &    0.25  \\
\multicolumn{1}{c|}{}        & \multicolumn{1}{c|}{}   & \multicolumn{1}{c|}{FFT-RadVLAD} &   11.45  &   0.06   &   0.16   &    0.20  \\
\multicolumn{1}{c|}{}        & \multicolumn{1}{c|}{\multirow{-5}{*}{\begin{tabular}[c]{@{}c@{}}Twins\\ Island\end{tabular}}}   & \multicolumn{1}{c|}{\cellcolor[HTML]{EFEFEF}Ours} & \cellcolor[HTML]{C4E1E6}\textbf{0.03}    & \cellcolor[HTML]{C4E1E6}\textbf{0.84}  & \cellcolor[HTML]{C4E1E6}\textbf{0.74}   & \cellcolor[HTML]{C4E1E6}\textbf{0.92}    \\ \cline{2-7} 
\multicolumn{1}{c|}{}        & \multicolumn{1}{c|}{}   & \multicolumn{1}{c|}{RingKey}     &     0.31     &   \underline{0.78}   &   \underline{0.72}   &   \underline{0.80}  \\
\multicolumn{1}{c|}{}        & \multicolumn{1}{c|}{}   & \multicolumn{1}{c|}{Referee}     &     \underline{0.21}     &   0.67   &   0.69   &   0.72  \\
\multicolumn{1}{c|}{}        & \multicolumn{1}{c|}{}   & \multicolumn{1}{c|}{RadVLAD}     &     31.85    &   0.07   &   0.19   &   0.34  \\
\multicolumn{1}{c|}{}        & \multicolumn{1}{c|}{}   & \multicolumn{1}{c|}{FFT-RadVLAD} &     33.13    &   0.02   &   0.06   &   0.17  \\
\multicolumn{1}{c|}{\multirow{-15}{*}{MOANA}} & \multicolumn{1}{c|}{\multirow{-5}{*}{\begin{tabular}[c]{@{}c@{}}Complex\\ Island\end{tabular}}} & \multicolumn{1}{c|}{\cellcolor[HTML]{EFEFEF}Ours} & \cellcolor[HTML]{C4E1E6}\textbf{0.09}   & \cellcolor[HTML]{C4E1E6}\textbf{0.88}  & \cellcolor[HTML]{C4E1E6}\textbf{0.79}  & \cellcolor[HTML]{C4E1E6}\textbf{0.88}    \\ \hline
\multicolumn{2}{c|}{}               & \multicolumn{1}{c|}{RingKey}     &  0.23       & \underline{0.74}   &  \underline{0.67}  &  \textbf{0.91}      \\
\multicolumn{2}{c|}{}               & \multicolumn{1}{c|}{Referee}     &  \underline{0.21}       & 0.51   &  0.46  & 0.74        \\
\multicolumn{2}{c|}{}               & \multicolumn{1}{c|}{RadVLAD}     &  15.44      & 0.08   &  0.18  & 0.32        \\
\multicolumn{2}{c|}{}               & \multicolumn{1}{c|}{FFT-RadVLAD} &  17.37      & 0.05   &  0.13  & 0.28       \\
\multicolumn{2}{c|}{\multirow{-5}{*}{Own Coastal}}     & \multicolumn{1}{c|}{\cellcolor[HTML]{EFEFEF}Ours} & {\cellcolor[HTML]{C4E1E6}\textbf{0.12}} & {\cellcolor[HTML]{C4E1E6}\textbf{0.77}} & {\cellcolor[HTML]{C4E1E6}\textbf{0.68}} & {\cellcolor[HTML]{EBFFD8}\underline{0.89}} \\ \hline
\end{tabular}
}
\caption{Intra-session PR (Bold: Best, Underline: Runner-up)}\label{tab:pr_result}
\end{table}

\begin{table}[t!]
\centering
\resizebox{\columnwidth}{!}{
\begin{tabular}{cccccccc}
\hline
Target  & Query              & Method              & AUC     & F1      & ROC    & Recall@1             \\ \hline
\multicolumn{1}{c|}{}  & \multicolumn{1}{c|}{}               & \multicolumn{1}{c|}{RingKey}              &  0.87   &   0.82  &   0.69 & \underline{0.70} \\
\multicolumn{1}{c|}{}  & \multicolumn{1}{c|}{}               & \multicolumn{1}{c|}{Referee}              &  \underline{0.91}   &   \textbf{0.88}  &   \textbf{0.81} & 0.70 \\
\multicolumn{1}{c|}{}  & \multicolumn{1}{c|}{}               & \multicolumn{1}{c|}{RadVLAD}              &  0.08   &   0.22  &    0.27  & 0.13  \\
\multicolumn{1}{c|}{}  & \multicolumn{1}{c|}{}               & \multicolumn{1}{c|}{FFT-RadVLAD}          &  0.06   &   0.14  &    0.37 & 0.15 \\
\multicolumn{1}{c|}{}  & \multicolumn{1}{c|}{\multirow{-5}{*}{\begin{tabular}[c]{@{}c@{}}MOANA\\ Single\\ Island\end{tabular}}}  & \multicolumn{1}{c|}{\cellcolor[HTML]{EFEFEF}Ours} & \cellcolor[HTML]{C4E1E6}\textbf{0.92} & \cellcolor[HTML]{EBFFD8}\underline{0.85} & \cellcolor[HTML]{EBFFD8}\underline{0.76} & \cellcolor[HTML]{C4E1E6}\textbf{0.74}\\ \cline{2-7} 
\multicolumn{1}{c|}{}  & \multicolumn{1}{c|}{}               & \multicolumn{1}{c|}{RingKey}              &   \underline{0.86}   &  \underline{0.87}  &   \underline{0.61}   & \underline{0.77} \\
\multicolumn{1}{c|}{}  & \multicolumn{1}{c|}{}               & \multicolumn{1}{c|}{Referee}              &   0.57   &  0.53  &   0.31  & 0.36 \\
\multicolumn{1}{c|}{}  & \multicolumn{1}{c|}{}               & \multicolumn{1}{c|}{RadVLAD}              &   0.15   &  0.32  &     0.37  & 0.15 \\
\multicolumn{1}{c|}{}  & \multicolumn{1}{c|}{}               & \multicolumn{1}{c|}{FFT-RadVLAD}          &   0.05   &  0.16  &     0.13  & 0.11 \\
\multicolumn{1}{c|}{}  & \multicolumn{1}{c|}{\multirow{-5}{*}{\begin{tabular}[c]{@{}c@{}}MOANA\\ Twins\\ Island\end{tabular}}}   & \multicolumn{1}{c|}{\cellcolor[HTML]{EFEFEF}Ours} & \cellcolor[HTML]{C4E1E6}\textbf{0.95} & \cellcolor[HTML]{C4E1E6}\textbf{0.90} & \cellcolor[HTML]{C4E1E6}\textbf{0.77} & \cellcolor[HTML]{C4E1E6}\textbf{0.81}\\ \cline{2-7} 
\multicolumn{1}{c|}{}  & \multicolumn{1}{c|}{}               & \multicolumn{1}{c|}{RingKey}              &   \underline{0.73}  &  0.66   &  0.72  & \underline{0.44}\\
\multicolumn{1}{c|}{}  & \multicolumn{1}{c|}{}               & \multicolumn{1}{c|}{Referee}              &   0.69  &  \textbf{0.70}   &  \underline{0.75}   & 0.42\\
\multicolumn{1}{c|}{}  & \multicolumn{1}{c|}{}               & \multicolumn{1}{c|}{RadVLAD}              &   0.03  &     0.10    &       0.20  & - \\
\multicolumn{1}{c|}{}  & \multicolumn{1}{c|}{}               & \multicolumn{1}{c|}{FFT-RadVLAD}          &   0.02  &  0.06   &     0.14 & 0.06 \\
\multicolumn{1}{c|}{} & \multicolumn{1}{c|}{\multirow{-5}{*}{\begin{tabular}[c]{@{}c@{}}MOANA\\ Complex\\ Island\end{tabular}}} & \multicolumn{1}{c|}{\cellcolor[HTML]{EFEFEF}Ours} & \cellcolor[HTML]{C4E1E6}\textbf{0.80} & \cellcolor[HTML]{EBFFD8}\underline{0.69} & \cellcolor[HTML]{C4E1E6}\textbf{0.78} & \cellcolor[HTML]{C4E1E6}\textbf{0.45}\\ 
\cline{2-7} 
\multicolumn{1}{c|}{}  & \multicolumn{1}{c|}{}               & \multicolumn{1}{c|}{RingKey}              &  0.69   &  \underline{0.77}  &  0.44   & \underline{0.62}\\
\multicolumn{1}{c|}{}  & \multicolumn{1}{c|}{}               & \multicolumn{1}{c|}{Referee}              &  \underline{0.73}   &  0.70  &  \textbf{0.66}   & 0.48\\
\multicolumn{1}{c|}{}  & \multicolumn{1}{c|}{}               & \multicolumn{1}{c|}{RadVLAD}              &  0.09   &  0.21  &     0.38& 0.07\\
\multicolumn{1}{c|}{}  & \multicolumn{1}{c|}{}               & \multicolumn{1}{c|}{FFT-RadVLAD}          &  0.06   &  0.17  &     0.17  & 0.09\\
\multicolumn{1}{c|}{\multirow{-20}{*}{\begin{tabular}[c]{@{}c@{}}MOANA\\ Island\\ Expedition\end{tabular}}} & \multicolumn{1}{c|}{\multirow{-5}{*}{\begin{tabular}[c]{@{}c@{}}Own\\ Complex\\ Island\end{tabular}}} & \multicolumn{1}{c|}{\cellcolor[HTML]{EFEFEF}Ours}  & \cellcolor[HTML]{C4E1E6}\textbf{0.75} & \cellcolor[HTML]{C4E1E6}\textbf{0.79} & \cellcolor[HTML]{EBFFD8}\underline{0.53} & \cellcolor[HTML]{C4E1E6}\textbf{0.65}\\ \hline
\multicolumn{1}{c|}{}  & \multicolumn{1}{c|}{}               & \multicolumn{1}{c|}{RingKey}              &   \underline{0.87}  &   \underline{0.84}  &   \underline{0.70}    &  \textbf{0.70} \\
\multicolumn{1}{c|}{}  & \multicolumn{1}{c|}{}               & \multicolumn{1}{c|}{Referee}              &   0.77  &   0.73  &   0.65    &  0.53 \\
\multicolumn{1}{c|}{}  & \multicolumn{1}{c|}{}               & \multicolumn{1}{c|}{RadVLAD}              &   0.11  &   0.24  &     0.38   &  0.03\\
\multicolumn{1}{c|}{}  & \multicolumn{1}{c|}{}               & \multicolumn{1}{c|}{FFT-RadVLAD}          &   0.05  &   0.17  &     0.01   &  0.03\\
\multicolumn{1}{c|}{\multirow{-5}{*}{\begin{tabular}[c]{@{}c@{}}MOANA\\ Complex\\ Island\end{tabular}}}              & \multicolumn{1}{c|}{\multirow{-5}{*}{\begin{tabular}[c]{@{}c@{}}Own\\ Complex\\ Island\end{tabular}}}     & \multicolumn{1}{c|}{\cellcolor[HTML]{EFEFEF}Ours}  & \cellcolor[HTML]{C4E1E6}\textbf{0.91} & \cellcolor[HTML]{C4E1E6}\textbf{0.87} & \cellcolor[HTML]{C4E1E6}\textbf{0.82} & \cellcolor[HTML]{EBFFD8}\underline{0.69}\\ \hline
\multicolumn{1}{c|}{}  & \multicolumn{1}{c|}{}               & \multicolumn{1}{c|}{RingKey}              &   0.94  &   0.98  &   0.04  & 0.96\\
\multicolumn{1}{c|}{}  & \multicolumn{1}{c|}{}               & \multicolumn{1}{c|}{Referee}              &   \underline{0.97}  &   \textbf{0.99}  &   0.08  & \textbf{0.98}\\
\multicolumn{1}{c|}{}  & \multicolumn{1}{c|}{}               & \multicolumn{1}{c|}{RadVLAD}              & 0.03     &  0.10   &  \textbf{0.56}  & 0.02 \\
\multicolumn{1}{c|}{}  & \multicolumn{1}{c|}{}               & \multicolumn{1}{c|}{FFT-RadVLAD}          &   0.02      &  0.05       &   \underline{0.37}      & 0.02\\
\multicolumn{1}{c|}{}  & \multicolumn{1}{c|}{\multirow{-5}{*}{\begin{tabular}[c]{@{}c@{}}Pohang\\ Canal\\ 04\end{tabular}}}      & \multicolumn{1}{c|}{\cellcolor[HTML]{EFEFEF}Ours} & \cellcolor[HTML]{C4E1E6}\textbf{0.98} & \cellcolor[HTML]{EBFFD8}\underline{0.98} & \cellcolor[HTML]{EFEFEF}0.14 & \cellcolor[HTML]{EBFFD8}\underline{0.96}\\ \cline{2-7}
\multicolumn{1}{c|}{}  & \multicolumn{1}{c|}{}               & \multicolumn{1}{c|}{RingKey}              &   \underline{0.89}  &   \textbf{0.92}  &   \textbf{0.61}    & \textbf{0.85} \\
\multicolumn{1}{c|}{}  & \multicolumn{1}{c|}{}               & \multicolumn{1}{c|}{Referee}              &   0.89  &   0.90  &    0.58    & 0.82 \\
\multicolumn{1}{c|}{}  & \multicolumn{1}{c|}{}               & \multicolumn{1}{c|}{RadVLAD}              &   0.02      &  0.08       &  0.33       & 0.02 \\
\multicolumn{1}{c|}{}  & \multicolumn{1}{c|}{}               & \multicolumn{1}{c|}{FFT-RadVLAD}          &  0.02       &   0.08      &  \underline{0.58}       & 0.02\\
\multicolumn{1}{c|}{\multirow{-10}{*}{\begin{tabular}[c]{@{}c@{}}Pohang\\ Canal\\ 03\end{tabular}}}             & \multicolumn{1}{c|}{\multirow{-5}{*}{\begin{tabular}[c]{@{}c@{}}Pohang\\ Canal\\ 05\end{tabular}}}      & \multicolumn{1}{c|}{\cellcolor[HTML]{EFEFEF}Ours} & \cellcolor[HTML]{C4E1E6}\textbf{0.91} & \cellcolor[HTML]{EBFFD8}\underline{0.91} & \cellcolor[HTML]{EFEFEF}0.47 & \cellcolor[HTML]{EBFFD8}\underline{0.84}\\ \hline
\end{tabular}
}
\caption{Inter-session PR (Bold: Best, Underline: Runner-up)}\label{tab:gl_result}
\vspace{-4mm}
\end{table}

\subsection{Inter-session place recognition}
\begin{figure*}[!t]
    \centering
    \includegraphics[width=\linewidth]{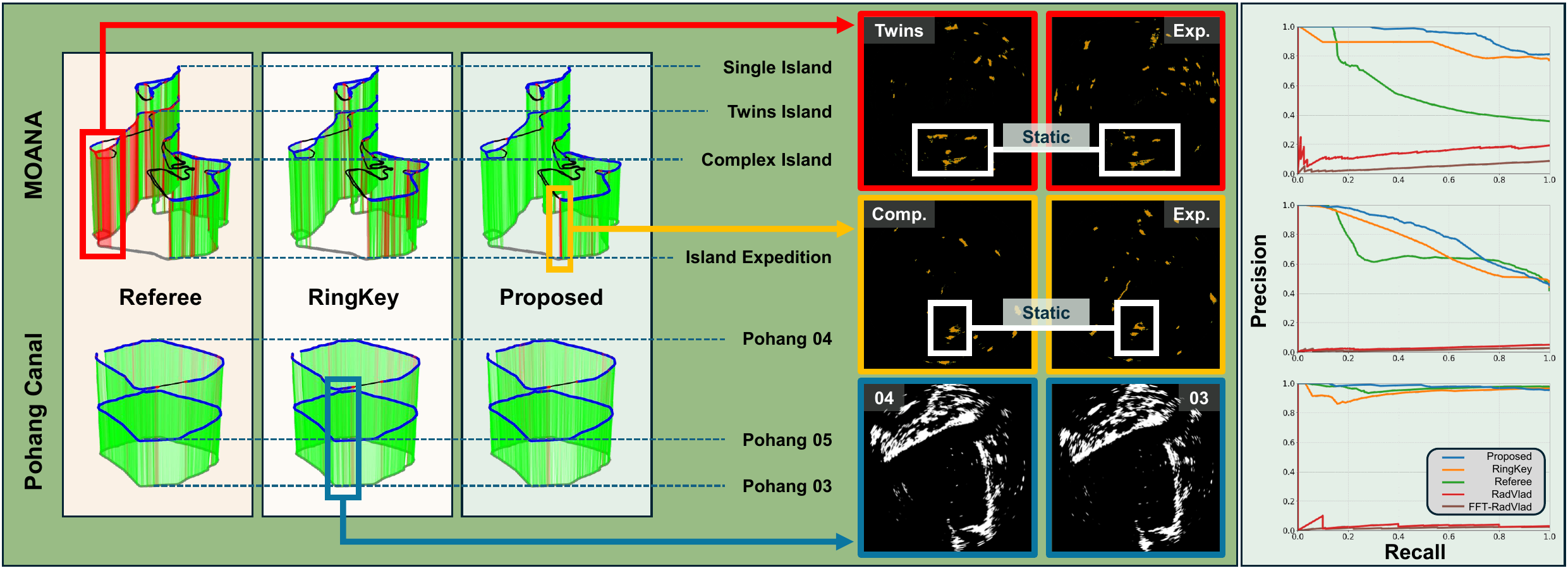}
    \caption{For the MOANA dataset, the performance differences among the algorithms are evident, with the proposed method demonstrating the highest robustness across all sequences. In contrast, the Pohang sequence yields reliable results for all algorithms. This discrepancy can be attributed to the presence of numerous dynamic objects in the MOANA dataset, whereas the Pohang Canal dataset features predominantly static scenes. These results highlight the robustness of the proposed algorithm in environments with dynamic elements. As illustrated in the right images, only the regions marked by white boxes correspond to static objects in the radar data. Despite this challenge, the proposed method successfully retrieves places in most cases (Red). However, in the most challenging scenario (Yellow), where static information is minimal, all algorithms failed to recognize the location as the same place.}
    \label{fig:gl_result}
    \vspace{-5mm}
\end{figure*}

For maritime vessels, global localization is a critical component, as GPS signals are frequently subject to interference and vessels often initiate operations from unknown locations. To ensure reliable initialization and enable loop closure for SLAM systems, we evaluate our algorithm with inter-session \ac{PR} performance. 
Similar to the results observed in intra-session \ac{PR}, the RadVLAD and FFT-RadVLAD descriptors were ineffective when applied to X-band radar data. In contrast, the other state-of-the-art algorithms demonstrated reliable performance. Notably, the proposed methods exhibited robustness across all sequences without performance degradation.
Detailed results are depicted in \tabref{tab:gl_result} and \figref{fig:gl_result}.

\begin{figure}[!t]
    \centering
    \includegraphics[trim= 0 0 0 50,clip,width=0.8\columnwidth]{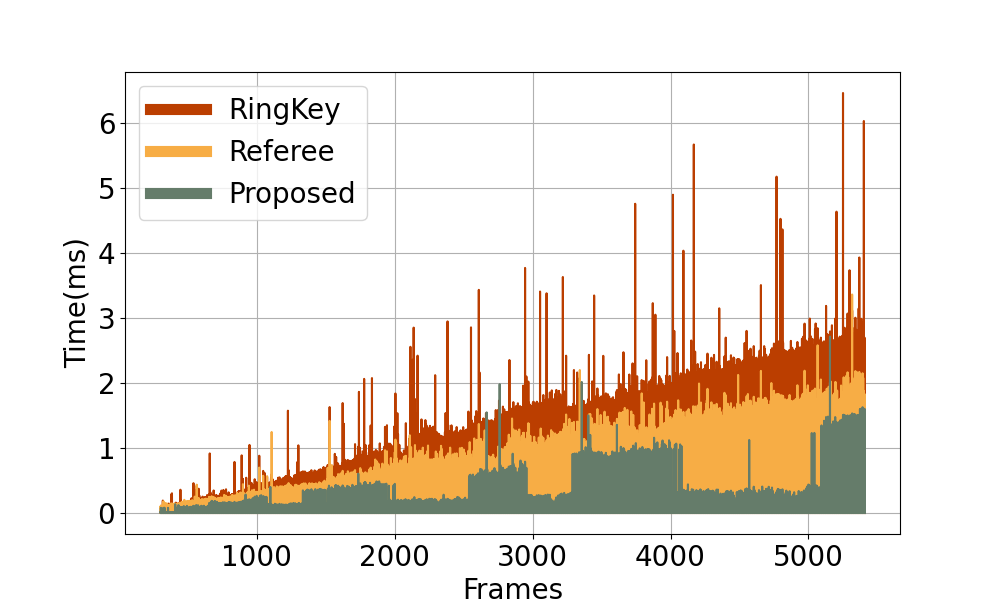}
    \caption{Computation time for descriptor retrieval is reduced in the proposed method by constraining the search space to candidates with similar cluster quantities.}
    \label{fig:time}
    \vspace{-6mm}
\end{figure}

\subsubsection{Pohang Canal Dataset}
The Pohang Canal Dataset consists of a structured, static environment with minimal dynamic elements, making it challenging to assess the failure case of the algorithms. As reflected in the results, all algorithms demonstrate strong performance across the sequences except for the VLAD descriptors. Furthermore, due to the identical route, the evaluation of the ROC curve is ineffective owing to the insufficient number of true negative instances.

\subsubsection{MOANA Dataset}
The results for inter-session \ac{PR} on the MOANA dataset are presented in the top side of \figref{fig:gl_result}. Compared to Referee, the proposed method demonstrates greater robustness on the Twins Island route, where ambiguous data is prevalent. Referee exhibits stable performance only on the simpler and more distinct dataset, Single Island. In comparison to RingKey, the proposed method achieves improved performance across all sequences.
For all algorithms, several common false negative cases were observed. This issue arises from the revisit distance criterion, where the $100m$ threshold is too large, causing post-corner poses that cannot be detected by the sensor to be included in the ground truth pool. To enable a more accurate evaluation of the algorithms, a manually curated or refined ground truth set is required, ensuring that both sensor observations share a common detectable area and fall within the defined distance threshold.

\subsubsection{Own Dataset}
A simple loop covering the MOANA Complex Island route was acquired, enabling the evaluation of inter-session \ac{PR} performance on both the MOANA Complex Island and Island Expedition sequences. Despite a time interval of approximately one year between the MOANA dataset and our data, and noticeable changes in the composition of anchored large vessels, the proposed algorithm robustly identified corresponding locations across both sequences.

\subsection{Computation Time}
The computation time of the algorithm is primarily determined by the size of the optimal descriptor. RadVLAD and FFT-RadVLAD require large descriptor sizes, resulting in increased computational costs. In contrast, the proposed method operates effectively with limited information, thereby reducing computation time. Furthermore, as depicted in \figref{fig:time}, applying thresholding based on cluster count minimizes unnecessary candidate searches. These characteristics make the proposed method particularly suitable for long-term navigation, which is common in maritime operations due to the typically extended duration of vessel routes.

\section{Ablation Study}
\label{sec:ablation}
% The robustness of the algorithm must be 
\begin{table*}[]
\centering
\vspace{-3mm}
\resizebox{\textwidth}{!}{
\begin{tabular}{cc|ccc|ccc|ccc}
\hline
\multirow{2}{*}{Criteria}                        & \multirow{2}{*}{Sequence} & \multicolumn{3}{c|}{Revisit Distance (Referee / RingKey / Proposed) {[}m{]}} & \multicolumn{3}{c|}{Descriptor Size} & \multicolumn{3}{c}{Cluster Threshold} \\ \cline{3-11} 
                                                 &                           & \hspace*{1mm}10\hspace*{1mm}  & \hspace*{1mm}50\hspace*{1mm}            & \hspace*{1mm}200 \hspace*{1mm}          & \hspace*{1mm}10\hspace*{1mm}         & \hspace*{1mm}20\hspace*{1mm}         & \hspace*{1mm}100\hspace*{1mm}       & \hspace*{1mm}1\hspace*{1mm}        & \hspace*{1mm}10\hspace*{1mm}        & \hspace*{1mm}100\hspace*{1mm}        \\ \hline
\multicolumn{1}{c|}{\multirow{3}{*}{AUC}}        & Single Island             &     0.18 / 0.17 / 0.15      &    0.63 / 0.76 / 0.82       &   0.98 / 0.88 / 0.92    &  0.88 &  0.94   &  0.95  & 0.86  &  0.96 &  0.96           \\
\multicolumn{1}{c|}{}                            & Twins Island              &     \;\;\;-\quad / 0.06 / 0.06      &    0.03 / 0.43 / 0.50       &   0.26 / 0.68 / 0.86       &  0.79 &  0.82   &  0.84  &  0.71 &  0.83 &  0.82           \\
\multicolumn{1}{c|}{}                            & Complex Island            &     0.06 / 0.06 / 0.03      &    0.54 / 0.62 / 0.63       &   0.69 / 0.86 / 0.92    &  0.72 &  0.86   &  0.88  &  0.80 &  0.89 &  0.88           \\ \hline
\multicolumn{1}{c|}{\multirow{3}{*}{F1-Score}}   & Single Island             &     0.32 / 0.36 / 0.40      &    0.78 / 0.73 / 0.85       &   0.93 / 0.81 / 0.82    &  0.81 &  0.89   &  0.90  &  0.76 &  0.89 &  0.90           \\
\multicolumn{1}{c|}{}                            & Twins Island              &     \;\;\;-\quad / 0.28 / 0.30       &    0.06 / 0.45 / 0.59       &   0.28 / 0.61 / 0.77      &  0.76 &  0.74   &  0.72  & 0.62  &  0.75 &  0.74           \\
\multicolumn{1}{c|}{}                            & Complex Island            &     0.12 / 0.14 / 0.13      &    0.59 / 0.56 / 0.59       &   0.70 / 0.79 / 0.82    &  0.67 &  0.77   &  0.78  & 0.71  &  0.79 &  0.78           \\ \hline
\multicolumn{1}{c|}{\multirow{3}{*}{ROC}} & Single Island             &     0.77 / 0.65 / 0.71      &    0.83 / 0.88 / 0.93       &   0.97 / 0.86 / 0.86    &  0.92 &  0.95   &  0.96  & 0.91  & 0.97  &  0.96           \\
\multicolumn{1}{c|}{}                            & Twins Island              &     \;\;\;-\quad / 0.62 / 0.77      &    0.53 / 0.77 / 0.87       &   0.71 / 0.83 / 0.91       &  0.91 &  0.93   &  0.93  & 0.87  & 0.94  &  0.92           \\
\multicolumn{1}{c|}{}                            & Complex Island            &     0.67 / 0.59 / 0.57      &    0.75 / 0.74 / 0.78       &   0.70 / 0.85 / 0.90    &  0.77 &  0.87   &  0.88  & 0.82  & 0.88  &  0.88           \\ \hline
\end{tabular}
}
\caption{Results for Ablation Study}\label{tab:abla_result}
\vspace{-5mm}
\end{table*}

\begin{figure}[!t]
    \vspace{-5mm}
    \centering
    \subfloat[$10\,m$]{
    \includegraphics[trim= 100 30 95 30,clip,width=0.22\columnwidth]{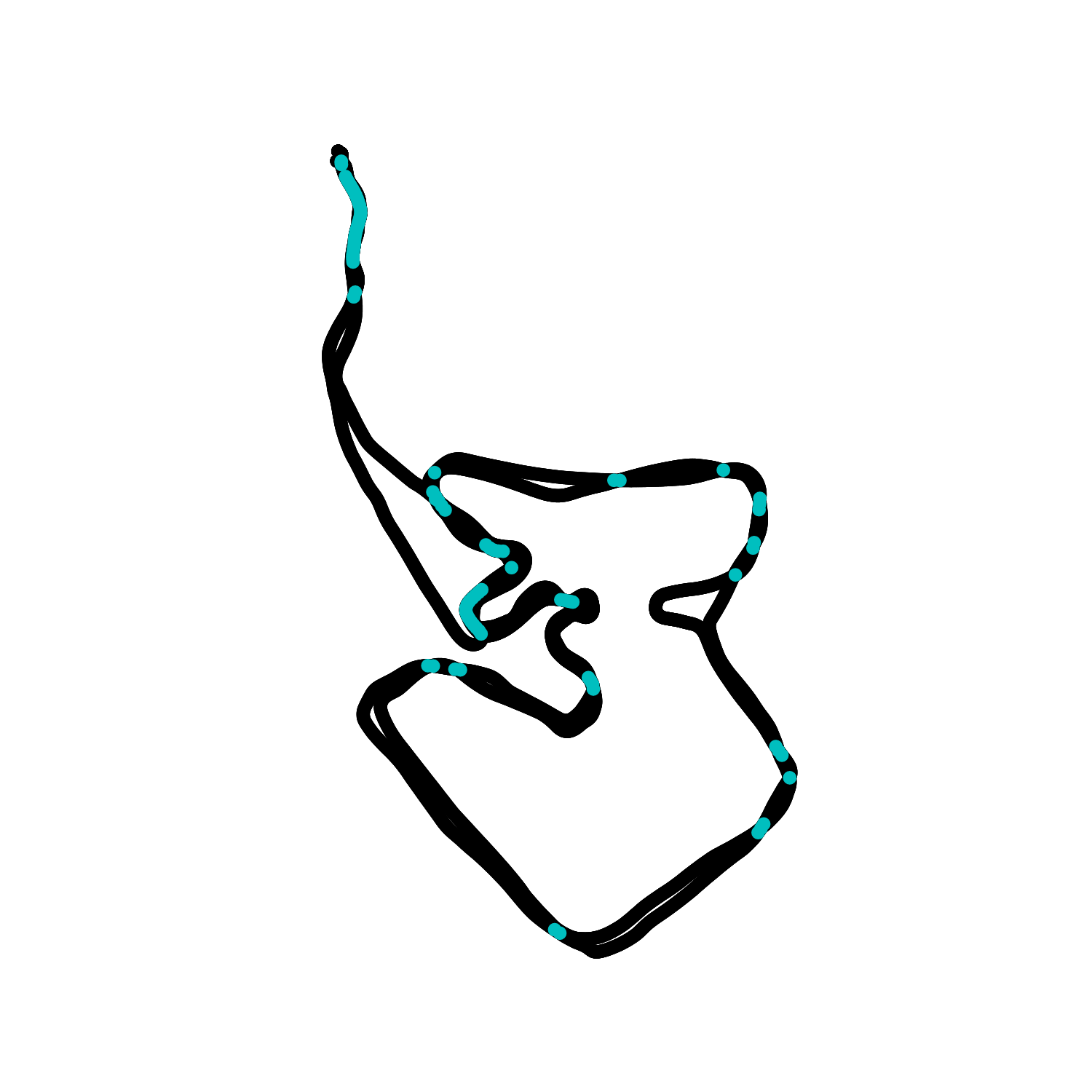}
    \label{fig:revisitness_10}
    }
    \subfloat[$50\,m$]{
    \includegraphics[trim= 100 30 95 30,clip,width=0.22\columnwidth]{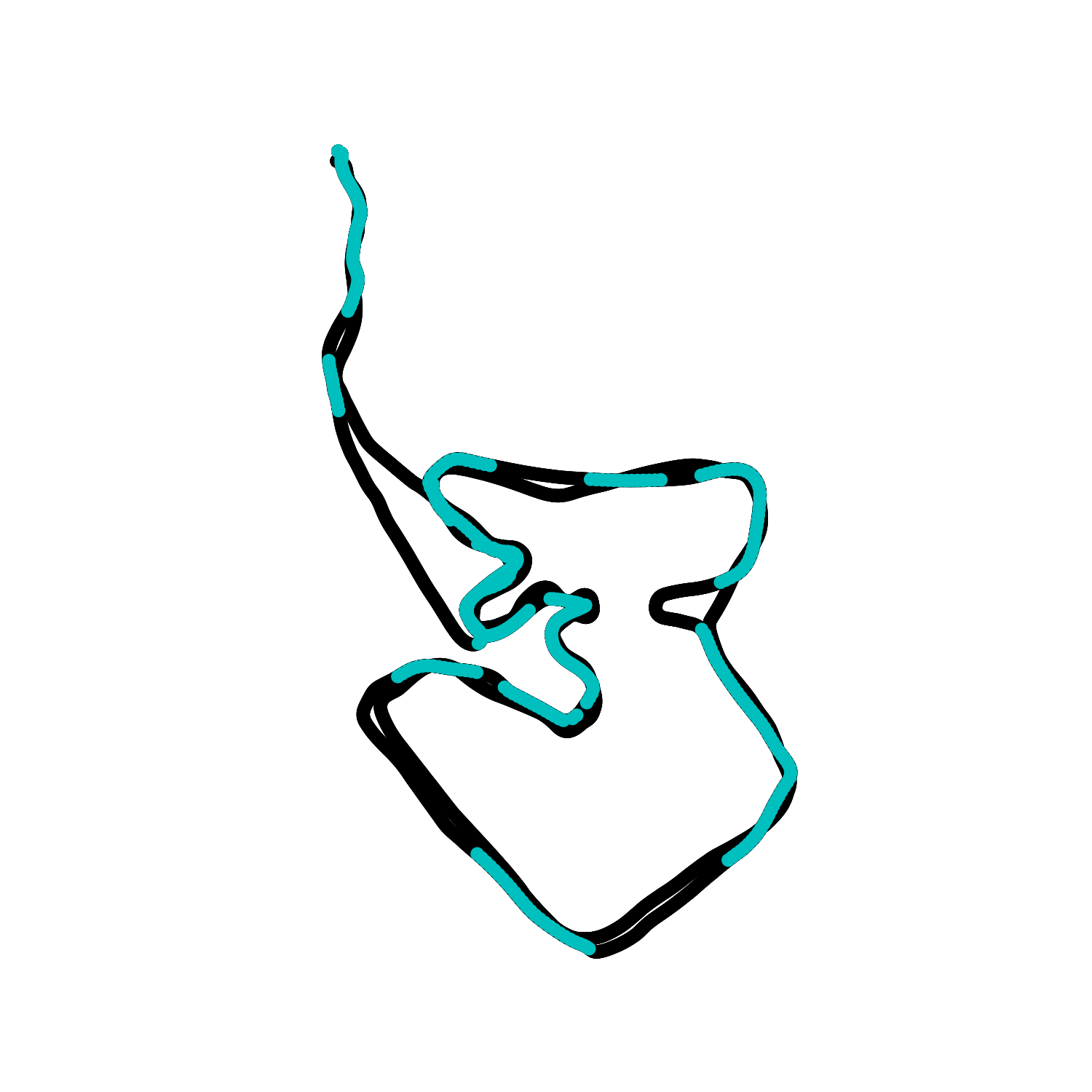}
    \label{fig:revisitness_50}
    }
    \subfloat[$100\,m$]{
    \includegraphics[trim= 100 30 95 30,clip,width=0.22\columnwidth]{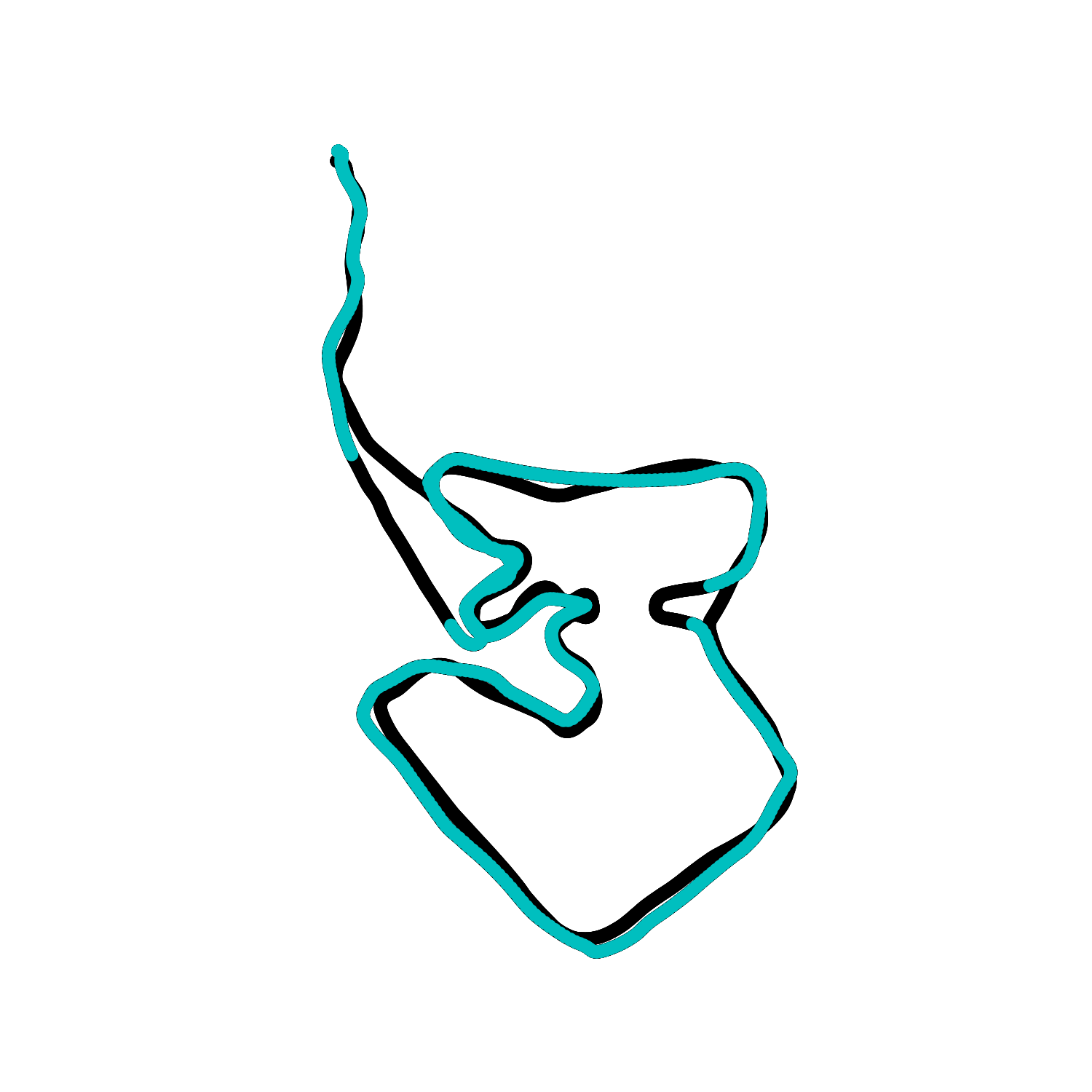}
    \label{fig:revisitness_100}
    }
    \subfloat[$200\,m$]{
    \includegraphics[trim= 100 30 95 30,clip,width=0.22\columnwidth]{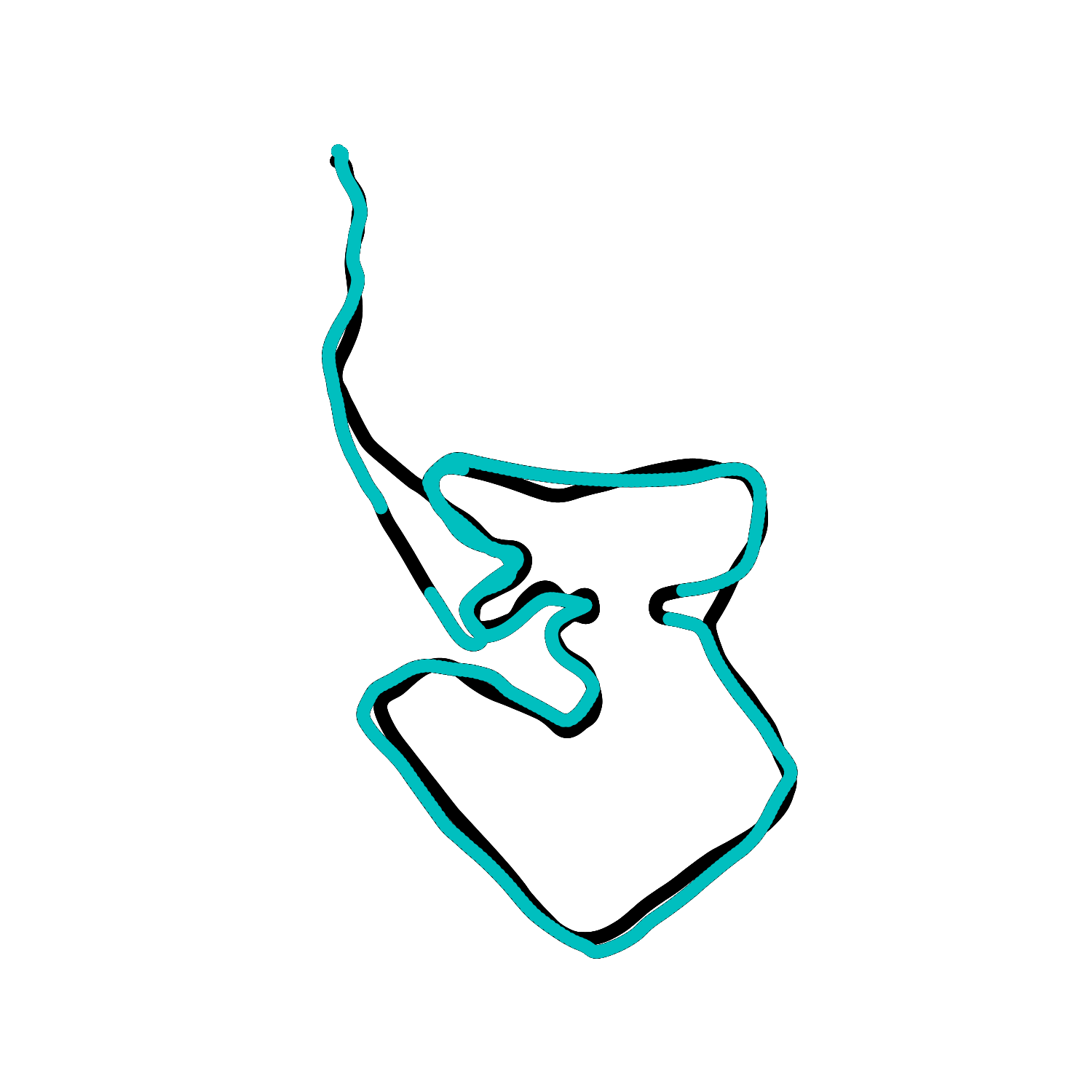}    \label{fig:revisitness_200}
    }
    \caption{Revisitness map corresponding to different revisit distance thresholds. Cyan-colored points indicate the locations where the vessel revisited. Due to the absence of defined lanes in maritime, maintaining identical routes is challenging. A $100\,m$ distance threshold is considered reasonable for evaluating \ac{PR} performance.}
    \label{fig:revisit_thres}
    \vspace{-2mm}
\end{figure}

\subsection{Revisit Distance Threshold}
There are no strict lanes for vessels, which results in potentially large distances between revisited locations. As shown in \figref{fig:revisit_thres}, a $10\,m$ revisit threshold does not yield sufficient ground truth data. This indicates that loop detection in maritime navigation should account for both sensor limitations and environmental constraints.
Considering both sensor and environmental conditions, an ablation study was conducted using distance thresholds of $10\,m$, $50\,m$, and $200\,m$. As anticipated, the $10\,m$ threshold caused insufficient ground truth resulting in low detection accuracy across all algorithms. Increasing the threshold to $50\,m$ produced more reliable results; however, it remained inadequate for fully covering ambiguous regions that fall below the minimum detection range of the X-band radar. At a $200\,m$ threshold, the results appeared more reasonable compared to smaller thresholds, but the inclusion of false ground truth led to a negative impact on \ac{PR} accuracy.
From the results of this ablation study, the lateral uncertainty of the X-band radar can be estimated based on the relationship between the distance threshold and \ac{PR} performance.

\subsection{Descriptor Size}
We employed the azimuth-wise integration of the ellipse polar image as the descriptor, where the vector size corresponds to the resolution of the range-wise information and directly influences the algorithm's computational cost. By varying the descriptor size from 10 to 100, we evaluated the performance of the algorithm. The results indicate that smaller descriptor lengths degrade performance, while beyond a certain size, further increases do not significantly affect the algorithm's effectiveness. To minimize computational complexity, catching an optimal descriptor size is recommended.

\subsection{Cluster Thresholding}
Cluster thresholding is applied to mitigate the influence of dynamic objects appearing in the same location. A threshold value of 1 allows only a single change relative to the base image, while a value of 100 includes nearly all images as candidates. Similar to the descriptor size results, thresholding has minimal impact at larger values but constrains performance at lower thresholds. 
In this context, determining the optimal threshold is necessary; however, the optimal value varies depending on the characteristics of the environment. An adaptive strategy, where the threshold is increased in regions with frequent dynamic object occurrences and decreased in more static environments, would provide an effective approach.

% \subsection{Range Resolution}
% X-band radar offers multiple resolution settings, enabling the acquisition of both high-resolution and low-resolution data. As the proposed method simplifies the entire detected clusters, it demonstrates robustness in low-resolution scenarios where state-of-the-art radar place recognition methods typically experience performance degradation. In contrast, existing methods perform reliably under high-resolution conditions. These results indicate that the proposed algorithm is well-suited for large-scale, low-resolution radar data.
% \begin{figure}[!t]
%     \centering
%     \includegraphics[width=\columnwidth]{./figs/dummy.png}
%     \caption{Multi-res x-band radar.}
%     \label{fig:rangeres}
%     \vspace{-2mm}
% \end{figure}
\section{Conclusion}
\label{sec:conclusion}
We proposed a place recognition method tailored for X-band radar, aiming to facilitate the initial stage of autonomous vessel navigation by perceiving the maritime environment. In contrast to conventional radar-based \ac{PR} approaches, the proposed method addresses the unique characteristics of X-band radar by employing an algorithm designed to suppress the influences of dynamic elements and simplify radar detections with elliptical clusters. \revision{In future work, we plan to investigate the challenges induced by vessel perturbations and examine the impact of sensor settings on X-band radar. This will be integrated into our investigation of radar-only SLAM and multi-sensor fusion approaches for robust maritime navigation.}
% Given the vast scale of the maritime environment, hierarchical place recognition is essential for achieving computationally efficient navigation performance. Assuming the multi-sensor framework, X-band radar estimates vessel location at a decameter-level resolution, while W-band radar achieves meter-level accuracy through structure-based localization. Subsequently, LiDAR and vision-based methods enable further refinement of the vessel’s pose via local feature alignment. As described in the example of future applications, X-band radar-based algorithms are expected to serve as a foundational component for maritime navigation systems.

% \newpage
% \newpage

%\section*{ACKNOWLEDGMENT}
\balance
\small
\bibliographystyle{IEEEtranN} %citeauthor
\bibliography{string-short,references}

\end{document}